\documentclass{article}
\usepackage{ijcai24}

\usepackage{times}
\usepackage{soul}
\usepackage{url}
\usepackage[hidelinks]{hyperref}
\usepackage[utf8]{inputenc}
\usepackage[small]{caption}
\usepackage{graphicx}
\usepackage{amsmath}
\usepackage{amsthm}
\usepackage{booktabs}
\usepackage{algorithm}
\usepackage{algorithmic}
\usepackage[switch]{lineno}
\usepackage{subcaption}

\usepackage{bbding}
\usepackage{bm}
\usepackage{multirow}
\usepackage{multicol}
\usepackage{lineno}
\usepackage{booktabs}
\usepackage{caption}
\usepackage{epsfig}
\usepackage{amssymb}
\usepackage{cuted}

\usepackage[table]{xcolor}

\title{Zero-shot High-fidelity and Pose-controllable Character Animation}

\author{
Bingwen Zhu$^{1,2}$
\and
Fanyi Wang$^{3*}$\and
Tianyi Lu$^{1,2}$\and
Peng Liu$^3$\and
Jingwen Su$^3$\and 
Jinxiu Liu$^4$\and \\
Yanhao Zhang$^3$\and
Zuxuan Wu$^{1,2}$\footnote{Corresponding Authors.}\and
Guo-Jun Qi$^5$\And
Yu-Gang Jiang$^{1,2}$
\\
\affiliations
{$^1$Shanghai Key Lab of Intell. Info. Processing, School of CS, Fudan University} \\
{$^2$Shanghai Collaborative Innovation Center of Intelligent Visual Computing} \\
$^3$OPPO AI Center\\
$^4$South China University Of Technology\\
$^5$Westlake University\\
}

\begin{document}

\maketitle

\begin{abstract}
     Image-to-video (I2V) generation aims to create a video sequence from a single image, which requires high temporal coherence and visual fidelity.     
     However, existing approaches suffer from inconsistency of character appearances and poor preservation of fine details. Moreover, they require a large amount of video data for training, which can be computationally demanding.
     To address these limitations, we propose PoseAnimate, a novel zero-shot I2V framework for character animation.
     PoseAnimate contains three key components: 1) a Pose-Aware Control Module (PACM) that incorporates diverse pose signals into text embeddings, to preserve character-independent content and maintain precise alignment of actions.
     2) a Dual Consistency Attention Module (DCAM) that enhances temporal consistency and retains character identity and intricate background details.
     3) a Mask-Guided Decoupling Module (MGDM) that refines distinct feature perception abilities, improving animation fidelity by decoupling the character and background.
     We also propose a Pose Alignment Transition Algorithm (PATA) to ensure smooth action transition.
     Extensive experiment results demonstrate that our approach outperforms the state-of-the-art training-based methods in terms of character consistency and detail fidelity. Moreover, it maintains a high level of temporal coherence throughout the generated animations.
\end{abstract}

\begin{figure*}[!t]
\centering
\includegraphics[width=\linewidth]{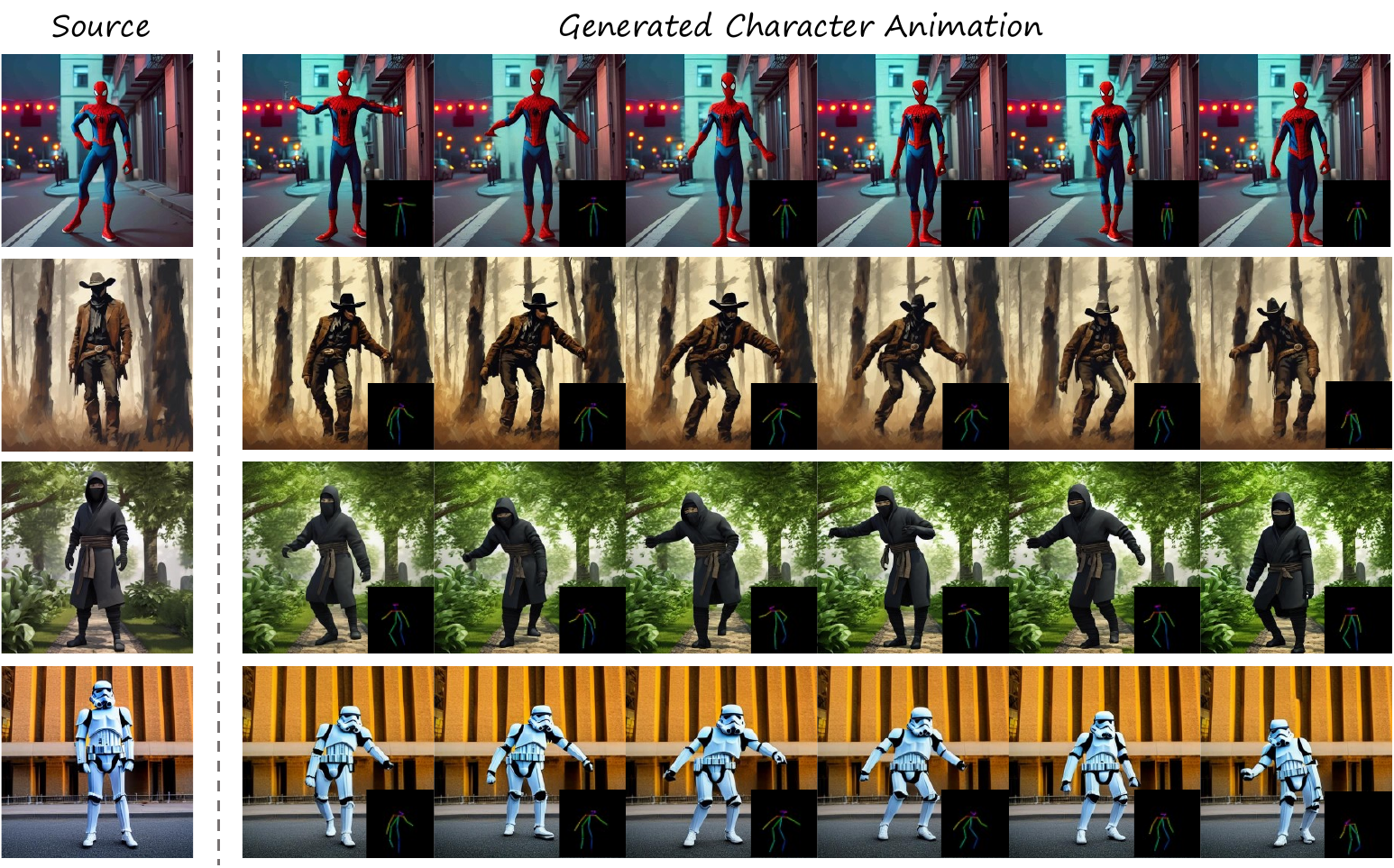} 
\caption{
PoseAnimate framework is capable of generating smooth and high-quality character animations for static character images across various pose sequences.}
\label{fig:head} 
\end{figure*}

\section{Introduction}
\label{sec:intro}
Image animation~\cite{siarohin2019first,siarohin2019animating,siarohin2021motion,wang2022latent,zhao2022thin} 
is a task that brings static images to life by seamlessly transforming them into dynamic and realistic videos. It involves the transformation of still images into a sequence of frames that exhibit smooth and coherent motions. In this task, character animation has gained significant attention due to its valuable applications in various scenarios, such as television production, game development, online retail and artistic creation, etc.  
However, minor motion variations hardly meet with the requirements.
The goal of character animation is to make the character in the image perform target pose sequences, while maintaining identity consistency and visual coherence. 
In early works, most of character animation was driven by traditional animation techniques, which involves meticulous frame-by-frame drawing or manipulation. In the subsequent era of deep learning, the advent of generative models~\cite{goodfellow2014generative,zhu2017unpaired,karras2019style} drove the shift towards data-driven and automated approaches~\cite{ren2020deep,chan2019everybody,zhang2022exploring}. However, there are still ongoing challenges in achieving highly realistic and visually consistent animations, especially when dealing with complex motions, fine-grained details, and long-term temporal coherence.

Recently, diffusion models~\cite{ho2020denoising} have demonstrated groundbreaking generative capabilities. Driven by the open source text-to-image diffusion model Stable Diffusion~\cite{rombach2022high}, the realm of video generation~\cite{xing2023survey} has achieved unprecedented progress in terms of visual quality and content richness. Hence, several endeavors~\cite{wang2023disco,xu2023magicanimate,hu2023animate} have sought to extrapolate the text-to-video (T2V) methods to image-to-video (I2V) by training additional image feature preserving networks and adapt them to the character animation task. 
Nevertheless, these training-based methods face challenges in accurately preserving features for arbitrary images and exhibit notable deficiencies in appearance control and loss of details. Additionally, they typically rely on extensive training data and significant computational overhead.

To this end, we contemplate employing a more refined and efficient approach, image reconstruction for feature preservation, to tackle this problem. We propose PoseAnimate, depicted in Fig.~\ref{fig:overview}, a zero-shot reconstruction-based I2V framework for pose controllable character animation.
PoseAnimate introduces a pose-aware control module (PACM), shown in Fig.~\ref{fig:embedding} which optimizes the text embedding twice based on the original and target pose conditions respectively, finally resulting a unique pose-aware embedding for each generated frame. This optimization strategy allows for the generated actions to be aligned with the target pose while keeping the character-independent scene consistent.
However, the introduction of a new target pose in the second optimization, which differs from the original pose, inevitably undermines the reconstruction of the character identity and background.
Thus, we further devise a dual consistency attention module (DCAM), as dedicated in the right part of Fig.~\ref{fig:overview}, to address the disruption, in addition to maintain a smooth temporal progression.
Since directly employing the entire attention map or key for attention fusion may result in loss of fine-grained detail perception.
We propose a mask-guided decoupling module (MGDM) to enable independent and focused spatial attention fusion for both the character and background. As such, our framework is able to capture the intricate character and background details, thereby effectively enhancing the fidelity of the animation.
In addition, for the sake of adaptation to various scales and positions of target pose sequences, a pose alignment transition algorithm (PATA) is designed to ensure pose alignment and smooth transitions.
Through combination of these novel modules, PoseAnimate achieves promising character animation results, as shown in Fig.~\ref{fig:head}, in a more efficient manner with lower computational overhead.

In summary, our contributions are as follows:
(1) We introduce a reconstruction-based approach to handle the task of character animation and propose PoseAnimate, a novel zero-shot framework, which generates coherent high-quality videos for arbitrary character images under various pose sequences, without any training of the network. To the best of our knowledge, we are the first to explore a training-free approach to character animation.
(2) We propose a pose-aware control module that enables precise alignment of actions while maintaining consistency across character-independent scenes.
(3) We decouple the character and the background regions, performing independent inter-frame attention fusion for them, which significantly enhances visual fidelity.
(4) Experiment results demonstrate the superiority of PoseAnimate compared with the state-of-the-art training-based methods in terms of character consistency and image fidelity.

\section{Related Work}
\label{sec:relate}
\subsection{Diffusion Models for Video Generation}
Image generation has made significant progress due to the advancement of Diffusion Models (DMs)~\cite{ho2020denoising}. Motivated by DM-based image generation~\cite{rombach2022high}, some works~\cite{yang2023diffusion,ho2022imagen,nikankin2022sinfusion,esser2023structure,xing2023simda,blattmann2023align,xing2023vidiff} explore DMs for video generation. Most video generation methods incorporate temporal modules to pre-trained image diffusion models, extending 2D U-Net to 3D U-Net. Recent works control the generation of videos with multiple conditions. For text-guided video generation, these works~\cite{he2022latent,ge2023preserve,gu2023reuse} usually tokenize text prompts with a pre-trained image-language model, such as CLIP~\cite{radford2021learning}, to control video generation through cross-attention. Due to the imperfect alignment between language and visual modalities in existing image-language models, text-guided video generation cannot achieve high textual alignment. Alternative methods~\cite{wang2023videocomposer,chen2023videocrafter1,blattmann2023stable} employ images as additional guidance for video generation. These works encode reference images to token space, helping capturing visual semantic information. VideoComposer~\cite{wang2023videocomposer} combines textual conditions, spatial conditions (e.g., depth, sketch, reference image) and temporal conditions (e.g., motion vector) through Spatio-Temporal Condition encoders. VideoCrafter1~\cite{chen2023videocrafter1} introduces a text-aligned rich image embedding to capture details both from text prompts and reference images. Stable Video Diffusion~\cite{blattmann2023stable} is a latent diffusion model for high-resolution T2V and I2V generation, which sets three different stages for training: text-to-image pretraining, video pretraining, and high-quality video finetuning.

\subsection{Video Generation with Human Pose}
 Generating videos with human pose is currently a popular task. Compared to other conditions, human pose can better guide the synthesis of motions in videos, which ensures good temporal consistency.  Follow your pose~\cite{ma2023follow} introduces a two-stage method to generate pose-controllable character videos.  
 Many studies~\cite{wang2023disco,karras2023dreampose,xu2023magicanimate,hu2023animate} try to generate character videos from still images via pose sequence, which needs to preserve consistency of appearance from source images as well.
 Inspired by ControlNet~\cite{zhang2023adding}, DisCo~\cite{wang2023disco} realizes disentangled control of human foreground, background and pose, which enables faithful human video generation.
To increase fidelity to reference human images, DreamPose~\cite{karras2023dreampose} proposes an adapter to models CLIP and VAE image embeddings.
MagicAnimate~\cite{xu2023magicanimate} adopts ControlNet~\cite{zhang2023adding} to extract motion conditions. It also introduces a appearance encoder to model reference images embedding.
Animate Anyone~\cite{hu2023animate} designs a ReferenceNet to extract detail features from reference images, combined with a pose guider to guarantee motion generation.

\section{Method}
\label{sec:method}

\begin{figure*}[!ht]
  \centering
  \includegraphics[width=\linewidth]{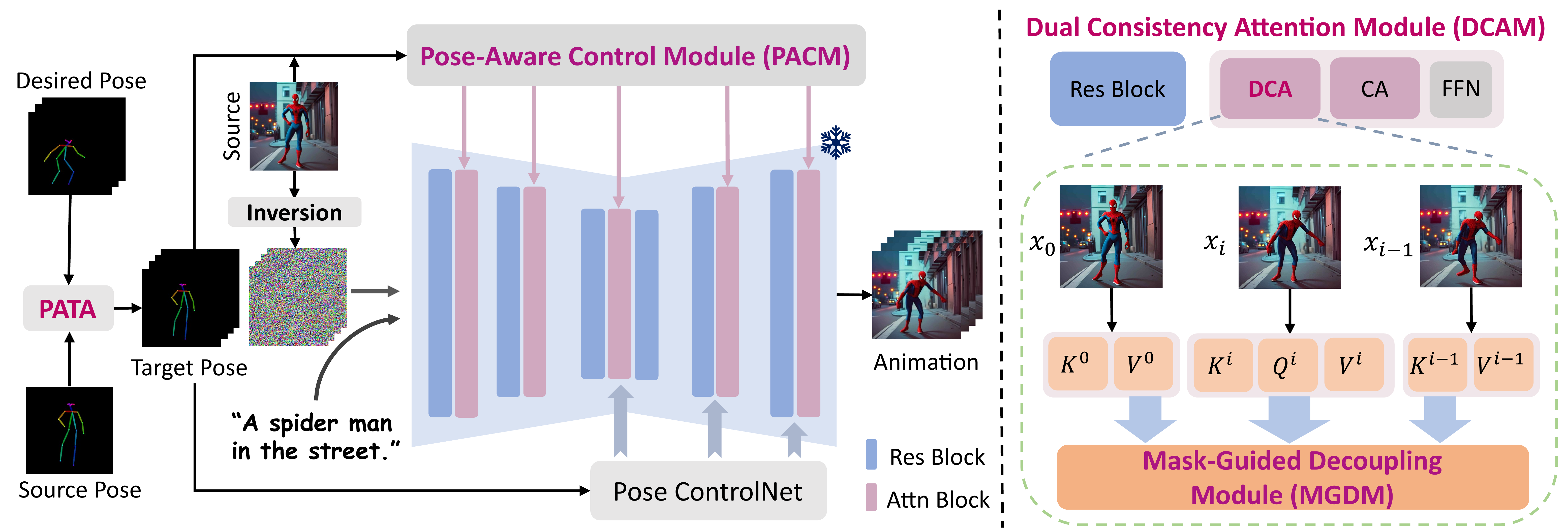} 
  \caption{\textbf{Overview of PoseAnimate.} The pipeline is on the \textbf{left}, we first utilize the Pose Alignment Transition Algorithm (PATA) to align the desired pose with a smooth transition to the target pose. We utilize the inversion noise of the source image as the starting point for generation. The optimized pose-aware embedding of PACM, in Sec.~\ref{sub:3.2}, serves as the unconditional embedding for input. The \textbf{right} side is the illustration of DCAM in Sec.~\ref{sub:3.3}. The attention block in this module consists of Dual Consistency Attention (DCA), Cross Attention (CA), and Feed-Forward Networks (FFN). Within DCA, we integrate MGDM to independently perform inter-frame attention fusion for the character and background, which further enhance the fidelity of fine-grained details.}
  \label{fig:overview}
\end{figure*}

Given a source character image $I_s$, and a desired pose sequence $P=\{p_i\}_{i=1}^{M}$, where $M$ is the length of the sequence. In the generated animation, we adopt a progressive approach to seamlessly transition the character from the source pose $p_s$ to the desired pose sequence $P=\{p_i\}_{i=1}^{M}$. We first facilitate the Pose Alignment Transition Algorithm (PATA) to smoothly interpolate $t$ intermediate frames between the source pose $p_s$ and the desired pose $P=\{p_i\}_{i=1}^{M}$. Simultaneously, it aligns each pose $p_i$ with the source pose $p_s$ to compensate for their discrepancies in terms of position and scale. As a result, the final target pose sequence is $P=\{p_i\}_{i=0}^{N}$, where $N=M+t$. It is worth noting that the first frame $x_0$ in our generated animation $X = \{x_i\}_{i=0}^{N}$ is identical to the source image $I_s$.
Secondly, we propose a pose-aware control module (PACM) that optimizes a unique pose-aware embedding for each generated frame.
This module can eliminate perturbation of the original character posture, thereby ensuring the generated actions aligned with the target pose $P$. 
Furthermore, it also maintains consistency of content irrelevant to characters. 
Thirdly, a dual consistency attention module (DCAM) is developed to ensure consistency of the character identity and improve temporal consistency. 
In addition, we design a mask-guided decoupling module (MGDM) to further enhance perception of character and background details. The overview of our PoseAnimate is depicted in Fig.~\ref{fig:overview}.

In this section, we begin with a brief introduction to Stable Diffusion in Sec.~\ref{sub:3.1}. 
Subsequently, Sec.~\ref{sub:3.2} introduces the incorporation of motion awareness into pose-aware embedding.
The proposed dual consistency control module is elaborated in Sec.~\ref{sub:3.3}, followed by the mask-guided decoupling module in Sec.~\ref{sub:3.4}.

\subsection{Preliminaries on Stable Diffusion} 
\label{sub:3.1}
Stable Diffusion~\cite{rombach2022high} has demonstrated strong text-to-image generation ability through a diffusion model in a latent space constructed by a pair of image encoder $\mathcal{E}$ and decoder $\mathcal{D}$. For an input image $\mathcal{I}$, the encoder $\mathcal{E}$ first maps it to a lower dimensional latent code $z_0 = \mathcal{E}(\mathcal{I})$, then Gaussian noise is gradually added to $z_0$ through the diffusion forward process:

\begin{equation} 
\label{eq:x_t-1}
q(\mathbf{z}_t|\mathbf{z}_{t-1}) = \mathcal{N}(\mathbf{z}_t; \sqrt{1 - \beta_t} \mathbf{z}_{t-1}, \beta_t\mathbf{I}), 
\end{equation}

\noindent where $t=1,...,T$, denotes the timesteps, $\beta_t \in (0,1)$ is a predefined noise schedule, and $\mathbf{I}$ is identity matrix. Through a parameterization trick, we can directly sample $z_t$ from $z_0$:

\begin{equation} 
\label{eq:x_0}
q(\mathbf{z}_t|\mathbf{z}_{0}) = \mathcal{N}(\mathbf{z}_t; \sqrt{\bar{\alpha_t}} \mathbf{z}_{0}, (1 - \bar{\alpha_t})\mathbf{I}),
\end{equation}

\noindent where $\bar{\alpha_t}=\prod_{i=1}^t \alpha_i$, and $\alpha_t= 1-\beta_t$. Diffusion models use a neural network $\epsilon_\theta$ to learn to predict the added noise $\epsilon$ by minimizing the mean square error of the predicted noise: 
\begin{equation} 
\label{eq:backward}
\min_\theta \mathbb{E}_{z,\epsilon \sim \mathcal{N}(0,I),t}[\Vert{\epsilon- \epsilon_\theta(z_t,t,\mathbf{c})}\Vert_2^2],
\end{equation}

\noindent where $\mathbf{c}$ is embedding of textual prompt. 
During inference, we can adopt a deterministic DDIM sampling~\cite{song2020denoising}, to iteratively recover a denoised representation $x_0$ from standard Gaussian noise $z_T, z_T \sim \mathcal{N}(0, I)$:

\begin{equation}
z_{t-1} = \sqrt{\bar\alpha_{t-1}} \underbrace{\hat{z}_{t \to 0}}_{\text{predicted } z_0} + \underbrace{\sqrt{1-\bar\alpha_{t-1}} \epsilon_\theta(z_t,t,\mathbf{c})}_{\text{direction pointing to } z_{t-1}},
\end{equation}

\noindent where $\hat{z}_{t \to 0}$ is the predicted $z_0$ at timestep $t$,

\begin{equation}
\hat{z}_{t \to 0} = \frac{z_t-\sqrt{1-\bar\alpha_t}\epsilon_\theta(z_t,t,\mathbf{c})}{\sqrt{\bar\alpha_t}}.
\end{equation}

Then $x_0$ is decoded into an output image $\mathcal{I}^{'} = \mathcal{D}(x_0)$ using the pre-trained decoder $\mathcal{D}$. 

\subsection{Pose-Aware Control Module}
\label{sub:3.2}
For generating a high fidelity character animation from a static image, two tasks need to be accomplished. Firstly, it is critical to preserve the consistency of original character and background in generated animation. 
In contrast to other approaches~\cite{karras2023dreampose,xu2023magicanimate,hu2023animate} that rely on training additional spatial preservation networks for consistency identity, we achieve it through a computationally efficient reconstruction-based method.
Secondly, the actions in the generated frames need to align with the target poses. Although the pre-trained OpenPose ControlNet~\cite{zhang2023adding} has great spatial control capabilities in controllable condition synthesis, our purpose is to discard the original pose and generate new continuous motion. Therefore, directly introducing pose signals through ControlNet may result in conflicts with the original pose, resulting in severe ghosting and blurring in motion areas.

\begin{figure}[!t]
  \centering
  \includegraphics[width=\columnwidth]{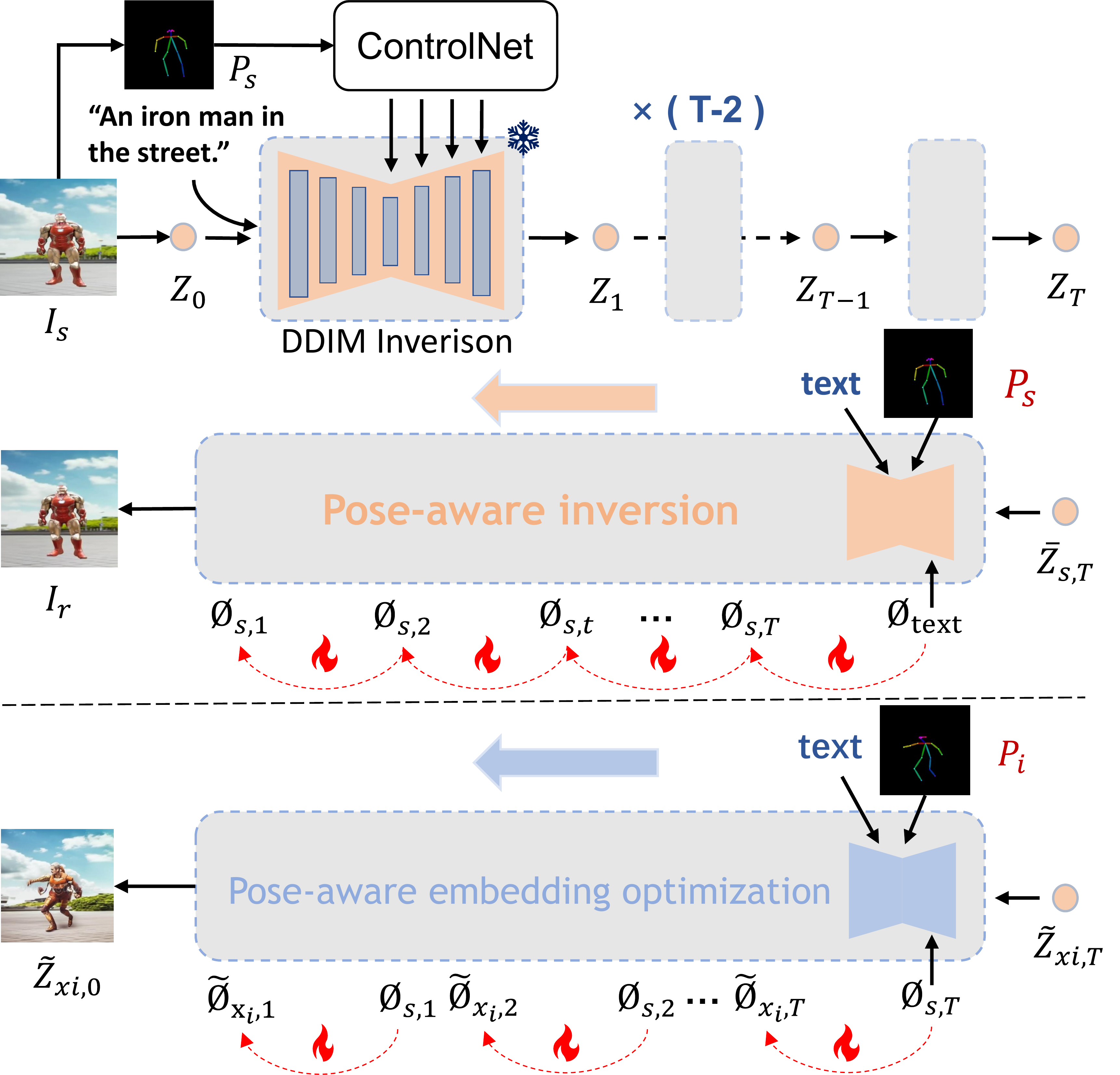}
  \caption{Illustration of Pose-Aware Control Module. Through two optimizations, the pose-aware embeddings are injected with motion awareness, which enables the alignment of generated actions with the target poses while maintaining consistency in character-independent scenes.}
  \label{fig:embedding}
\end{figure}

In light of this, we propose the pose-aware control module, as illustrated in Fig.~\ref{fig:embedding}. Inspired by the idea of inversion in image editing~\cite{mokady2023null}, we achieve the perception of pose signals by optimizing the text embedding $\varnothing_{text}$ twice based on the source pose $p_s$ and the target pose $p_i$, respectively. 
In the first optimization, i.e. pose-aware inversion, we involve a progressive optimization process for $\varnothing_{text}$ to accurately reconstruct the source image $I_s$ under the source pose $p_s$. We initialize $\bar{Z}_{s,T} = Z_T$, $\varnothing_{s,T} = \varnothing_{text}$, and perform the following optimization for the timesteps $t=T,\ldots,1$, each step for $n$ inner iterations:
\begin{equation}\label{eq}
\small
\min_{\varnothing_{s,t}} \Vert{Z_{t-1}-z_{t-1}(\bar{Z}_{s,t}, \varnothing_{s,t}, p_s,C)}\Vert_2^2,
\end{equation}
$z_{t-1}(\bullet)$ denotes applying DDIM sampling using latent code $\bar{Z}_{s,t}$, source embedding $\varnothing_{s,t}$, source pose $p_s$, and text prompt $C$. Building upon the optimized source embeddings $\{\varnothing_{s,t}\}_{t=1}^{T}$ obtained from this process, we then proceed with the second optimization, i.e. pose-aware embedding optimization, where we inject the target pose signals $P = \{p_i\}_{i=1}^{N}$ into the optimized pose-aware embeddings $\{\{\widetilde{\varnothing}_{x_i,t}\}_{t=1}^{T}\}_{i=1}^{N} $, as detailed in Alg.~\ref{alg:algorithm}. Perceiving the target pose signals, these optimized pose-aware embeddings $\{\{\widetilde{\varnothing}_{x_i,t}\}_{t=1}^{T}\}_{i=1}^{N} $ ensure a flawless alignment between the generated character actions and the target poses, while upholding the consistency of character-independent content.

Specifically, to incorporate the pose signals, we integrate ControlNet into all processes of the module. Diverging from null-text inversion~\cite{mokady2023null} that achieves image reconstruction by optimizing unconditional embeddings~\cite{ho2022classifier}, our pose-aware inversion optimizes the conditional embedding $\varnothing_{text}$ of the text prompt $C$ during the reconstruction process. The motivation stems from the observation that conditional embedding contains more abundant and robust semantic information, which endows it with a heightened potential for encoding pose signals.

\begin{algorithm}[tb]
\caption{Pose-aware embedding optimization.}
\label{alg:algorithm}
\textbf{Input}: Source character image $I_s$, source character pose $p_s$, text prompt $C$, and target pose sequence $P = \{p_i\}_{i=1}^{N}$, number of frames N, timestep T. \\
\textbf{Output}: Optimized source embeddings $\{\varnothing_{s,t}\}_{t=1}^{T}$, Optimized pose-aware embeddings $\{\{\widetilde{\varnothing}_{x_i,t}\}_{t=1}^{T}\}_{i=1}^{N} $ , and latent code $Z_T$ . \\
\begin{algorithmic}[1] 
    \STATE Set guidance scale = 1.0. Calculate DDIM inversion~\cite{dhariwal2021diffusion} latent code $Z_0,...,Z_T$ corresponding to input image $I_s$.
    \STATE Set guidance scale = 7.5. Obtain optimized source embeddings $\{\varnothing_{s,t}\}_{t=1}^{T}$ through pose-aware inversion.
    \FOR{$i = 1, 2, ..., N$}
        \STATE Initialize $\widetilde{Z}_{x_i,T} = Z_T$, $ \{\widetilde{\varnothing}_{x_i,t}\}_{t=1}^{T} = \{\varnothing_{s,t}\}_{t=1}^{T}$;
        \FOR{$t = T, T-1, ..., 1$}
            \STATE $\widetilde{Z}_{x_i,t-1}\leftarrow \text{Sample}(\widetilde{Z}_{x_i,t}, \epsilon_\theta(\widetilde{Z}_{x_i,t}, \widetilde{\varnothing}_{x_i,t}, p_i, C,t))$;
            \STATE $ \widetilde{\varnothing}_{x_i,t} \leftarrow  \widetilde{\varnothing}_{x_i,t} - \eta \nabla_{\widetilde{\varnothing}} \text{MSE}(Z_{t-1}, \widetilde{Z}_{x_i,t-1}) $;
        \ENDFOR
    \ENDFOR
    \STATE \textbf{Return} $Z_T, \{\varnothing_{s,t}\}_{t=1}^{T} , \{\{\widetilde{\varnothing}_{x_i,t}\}_{t=1}^{T}\}_{i=1}^{N} $
\end{algorithmic}
\end{algorithm}

\subsection{Dual Consistency Attention Module}
\label{sub:3.3}

Although the pose-aware control module accurately captures and injects body poses, it may unintentionally alter the identity of the character and the background details due to the introduction of different pose signals, as demonstrated by the example $\widetilde{Z}_{x_i,0}$ in Fig.~\ref{fig:embedding}, which is undesirable. Since self-attention layers in the U-Net~\cite{ronneberger2015u} play a crucial role in controlling appearance, shape, and fine-grained details, existing attention fusion paradigms commonly employ cross-frame attention mechanism~\cite{ni2022expanding}, to facilitate spatial information interaction across frames:
\begin{equation}
\label{eq:cross}
\text{{Attention}}(Q^i, K^j, V^j) = \text{softmax}\left(\frac{{Q^i(K^j)^\top}}{{\sqrt{d}}}\right) V^j ,
\end{equation}
\noindent where $Q^i$ is the query feature of frame $x_i$, and $K^j, V^j$ correspond to the key feature and value feature of frame $x_j$. As pose $p_1$ is identical to the source pose $p_s$,  the reconstruction of frame $x_0$ remains undisturbed, allowing for a perfect restoration of the source image $I_s$. Hence, we can compute the cross-frame attention between each subsequent frame $\{x_i\}_{i=1}^{N}$ with the frame $x_0$ to ensure the preservation of identity and intricate details. However, solely involving frame $x_0$ in the attention fusion would bias the generated actions towards the original action, resulting in ghosting artifacts and flickering. Consequently, we develop the Dual Consistency Attention Module (DCAM) by replacing self-attention layers with our dual consistency attention (DC Attention) to address the issue of appearance inconsistency and improve temporal consistency. The DC Attention mechanism operates for each subsequent frame $x_i$ as follows:
\begin{equation}
\begin{aligned}
\text{CFA}_{i,j} = \text{{Attention}}(Q^i, K^j, &V^j), \\
\text{Dual Consistency Attention}(x_i) :=&\text{DCA}_{i}= \\ \lambda_1 * \text{CFA}_{i,0}
 + \lambda_2 * \text{CFA}_{i,i-1} +  \lambda_3 & *  \text{CFA}_{i,i} , \\
\end{aligned}
\label{eq:attn}
\end{equation}
\noindent where $\text{CFA}_{i,j}$ refers to cross-frame attention between frames $x_i$ and $x_j$. $\lambda_1,\lambda_2,\lambda_3 \in (0,1)$ are hyper-parameters, and $\lambda_1 + \lambda_2 + \lambda_3 =1 $. They jointly control the participation of the initial frame $x_0$, the current frame $x_i$ and the preceding frame $x_{i-1}$ in the DC Attention calculation. In the experiment, we set $\lambda_1 = 0.7 $ and $\lambda_2 = \lambda_3 = 0.15$ to enable the frame $x_0$ to be more involved in the spatial correlation control of the current frame for the sake of better appearance preservation. Apart from this, retaining a relatively small portion of feature interaction for the current frame and the preceding frame simultaneously is promised to enhance motion stability and improve temporal coherence of the generated animation. 

Furthermore, it is vital to note that we do not replace all the U-Net~\cite{ronneberger2015u} transformer blocks with DCAM. We find that incorporating the DC Attention only in the upsampling blocks of the U-Net architecture while leaving the remaining unchanged allows us to maintain consistency with the identity and background details of the source, without compromising the current frame's pose and layout.

\subsection{Mask-Guided Decoupling Module}
\label{sub:3.4}
Directly utilizing the entire image features for attention fusion can result in a substantial loss of fine-grained details. To address this problem, we propose the mask-guided decoupling module, which decouples the character and background and enables individual inter-frame interaction to further refine spatial feature perception. 

For the source image $I_s$, we obtain a precise body mask $M_s$ (i.e. $M_{x_0}$) that separates the character from the background by an off-the-shelf segmentation model~\cite{liu2023lightweight}. The target pose prior is insufficient to derive body mask for each generated frame of the character. 
Considering the strong semantic alignment capability of cross attention layers mentioned in Prompt-to-prompt~\cite{hertz2022prompt}, we extract the corresponding body mask $M_{x_i}$ for each frame from the cross attention maps.
With $M_s$ and $M_{x_i}$, only attentions of character and background within corresponding region are calculated, according to the mask-guided decoupling module as follows:
\begin{equation}
\begin{aligned}
\label{eq:attn_mask}
\text{K}^c_{j}=M_{x_j} &\odot \text{K}_{j} ,  \text{K}^b_{j}=(1-M_{x_j}) \odot \text{K}_{j}, \\
\text{V}^c_{j}=M_{x_j} &\odot \text{V}_{j} ,  \text{V}^b_{j}=(1-M_{x_j}) \odot \text{V}_{j}, \\
\text{CFA}^c_{i,j} &= \text{Attention}(Q^i, {K}^c_{j}, {V}^c_{j}), \\
\text{CFA}^b_{i,j} &= \text{Attention}(Q^i, {K}^b_{j}, {V}^b_{j}),
\end{aligned}
\end{equation}
\noindent where $\text{CFA}^c_{i,j} $ is the attention output in character between frame $x_i$ and $x_j$,  and $\text{CFA}^b_{i,j} $ is for the background.
Then we can get the final DC Attention output:
\begin{equation}
\begin{aligned}
\label{eq:final_attn}
\text{DCA}^c_{i} &= \lambda_1 * \text{CFA}^c_{i,0} + \lambda_2 * \text{CFA}^c_{i,i-1} +  \lambda_3  *  \text{CFA}^c_{i,i}, \\
\text{DCA}^b_{i} &= \lambda_1 * \text{CFA}^b_{i,0} + \lambda_2 * \text{CFA}^b_{i,i-1} +  \lambda_3  *  \text{CFA}^b_{i,i}, \\
\text{DCA}_i &= M_{x_i} \odot \text{DCA}^c_{i} +(1-M_{x_i}) \odot  \text{DCA}^b_{i} ,
\end{aligned}
\end{equation}
\noindent for $i=1,...,N$. The proposed decoupling module introduces explicit learning boundary between the character and background, allowing the network to focus on their respective content independently rather than blending features. Consequently, the intricate details of both the character and background are preserved, leading to a substantial improvement in the fidelity of the animation.

\section{Experiment}

\subsection{Experiment Settings}
We implement PoseAnimate based on the public pre-trained weights of ControlNet~\cite{zhang2023adding} and Stable Diffusion~\cite{rombach2022high} v1.5. For each generated character animation, we generate $N = 16$ frames with a unified $512 \times 512$ resolution. In the experiment, we use DDIM sampler~\cite{song2020denoising} with the default hyperparameters: number of diffusion steps $T = 50$ and guidance scale $w = 7.5$. For the pose-aware control module, loss function of optimizing text embedding $\varnothing_{text}$ is MSE. The optimization iterations are $250$ in total with $n=5$ inner iterations per step, and the optimizer is Adam. All experiments are performed on a single NVIDIA A100 GPU.

\subsection{Comparison Result}
We compare our PoseAnimate with several state-of-the-art methods for character animation: \textbf{MagicAnimate}~\cite{xu2023magicanimate} and \textbf{Disco}~\cite{wang2023disco}.  
For MagicAnimate (MA), both DensePose~\cite{guler2018densepose} and OpenPose signals of the same motion are applied to evaluate the performances.
We leverage the official open source code of Disco to test its effectiveness. 
Additionally, we construct a competitive character animation baseline by IP-Adapter~\cite{ye2023ip} with ControlNet~\cite{zhang2023adding} and spatio-temporal attention~\cite{wu2023tune}, which is termed as \textbf{IP+CtrlN}.
It is worth noting that these methods are all training-based, while ours does not require training.

\paragraph{Qualitative Results.}
\begin{figure*}[!t]
  \centering
  \includegraphics[width=\linewidth]{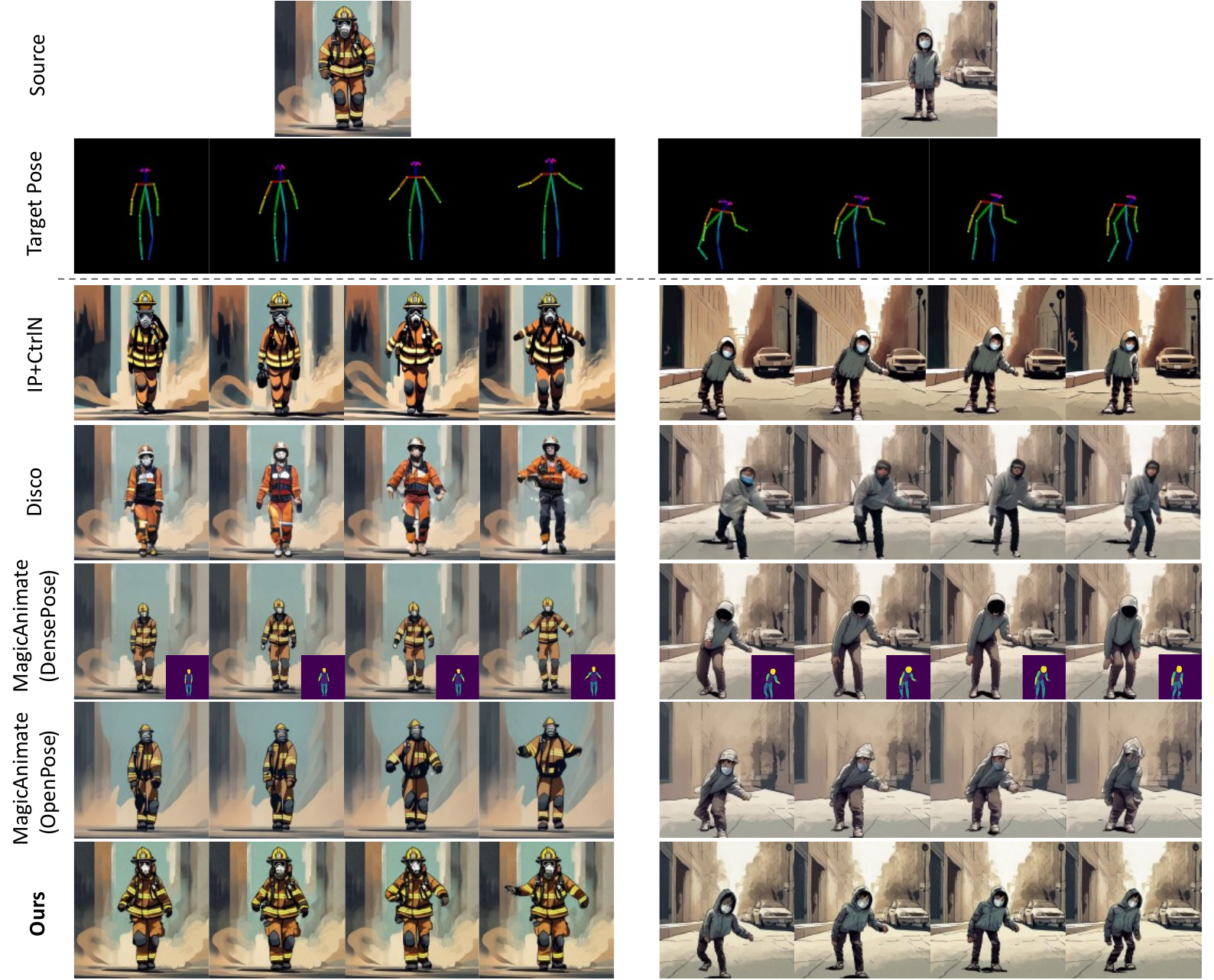}
  \caption{Qualitative comparison between our PoseAnimate and other training-based state-of-the-art character animation methods. We overlay the corresponding DensePose on the bottom right corner of the MagicAnimate (Densepose) synthesized frames. Previous methods suffer from inconsistent character appearance and details lost. Source prompt: ``A firefighters in the smoke."(left)``A boy in the street."(right).}
  \label{fig:quality}
\end{figure*}

We set up two different levels of pose for the experiments to fully demonstrate the superiority of our method. 
The visual comparison results are shown in Fig.~\ref{fig:quality}, with the left side displaying simple actions and the right side complex actions. 
Although IP+CtrlN has good performance on identity preservation, it fails to maintain details and inter-frame consistency. 
Disco completely loses character appearance, and severe frame jitter leads to ghosting shadows and visual collapse for complex actions.
MagicAnimate performs better than the other two methods, but it still encounters inconsistencies in character appearance at a more fine-grinded level guided by DensePose.
It is also unable to preserve background and character details accurately, e.g., vehicle textures and masks of the firefighter and the boy in Fig.~\ref{fig:quality}.
MagicAnimate under OpenPose signal conditions has worse performances than that under DensePose.
Our method exhibits the best performance on image fidelity to the source image and effectively preserves complex fine-grained appearance details and temporal consistency. 

\paragraph{Quantitative Results.}
\begin{table}[!t]
\centering
  \setlength{\tabcolsep}{0pt} 
  \begin{tabular*}{0.95\columnwidth}{@{\extracolsep{\fill}}lcccc@{}}
    \toprule
    \textbf{Method} &  LPIPS $\downarrow$  & CLIP-I $\uparrow$  & FC $\uparrow$ &  WE $\downarrow$ \\
    \midrule
    IP+CtrlN &  0.466  & 0.937  &  94.88 &   0.1323    \\
    Disco  &  0.278  &  0.811 &  92.23 &    0.0434   \\
    MA (DensePose) &  0.273  & 0.870  & \textbf{97.87}  &  \textbf{0.0193}   \\
    MA (OpenPose) &  0.411  & 0.867  &  97.63 &    0.0261    \\
    Ours &  \textbf{0.247}  &  \textbf{0.948} & 97.33  &  0.0384   \\
    \bottomrule
  \end{tabular*}
 \caption{Quantitative comparison between PoseAnimate and other training-based state-of-the-art methods. The best average performance is in bold. ↑ indicates higher metric value and represents better performance and vice versa. MA stands for MagicAnimate.}
\label{tab:quanti}
\end{table}
For quantitative analysis, we first randomly sample 50 in-the-wild image-text pairs and 10 different disered pose sequences to conduct evaluations. 
In this section, we adopt four evaluation metrics:
(1) LPIPS~\cite{zhang2018unreasonable} measures the fidelity between generated frames and the source image. (2) CLIP-I~\cite{ye2023ip} represents the similarity of CLIP~\cite{radford2021learning} image embedding between generated frames and the source image. (3) Frame Consistency (FC)~\cite{esser2023structure} evaluates video continuity by computing the average CLIP cosine similarity of two consecutive frames. 
(4) Warping Error (WE)~\cite{liu2023evalcrafter} evaluates the temporal consistency of the generated animation through the Optical Flow algorithm~\cite{teed2020raft}. 
Quantitative results are provided in Tab.~\ref{tab:quanti}.
Our method achieves the best scores on LPIPS and CLIP-I, and significantly surpasses other comparison methods in terms of fidelity to the source image, demonstrating outstanding detail preservation capability. 
In addition, PoseAnimate outperforms two training-based methods in terms of interframe consistency and obtains a good Warping Error score, illustrating that it is able to ensure good temporal coherence without additional training.

To further make a comprehensive quantitative performance comparison, we also follow the experimental settings in MagicAnimate, and evaluate both image fidelity and video quality on two benchmark datasets, namely TikTok~\cite{jafarian2021learning} and TED-talks~\cite{siarohin2021motion}. we compare FID-VID~\cite{balaji2019conditional} and FVD~\cite{unterthiner2018towards} metrics for video quality, as well as two essential image fidelity metrics, L1 and FID~\cite{heusel2017gans}. The experimental results are presented in Tab.~\ref{tab:comp}, where PoseAnimate achieves state-of-the-art image fidelity while maintaining competitive video quality.

\begin{table}[!t]
  \centering
    \begin{subtable}[t]{\columnwidth}
    \centering
  \begin{tabular}{lcccc}
    \toprule
    \multirow{2}{*}{Method~~}&\multicolumn{2}{c}{Image} &\multicolumn{2}{c}{Video}\\
    \cmidrule(r){2-3} \cmidrule(r){4-5} & L1\(\downarrow\)~~ & FID\(\downarrow\)~~ & FID-VID\(\downarrow\)~~& FVD\(\downarrow\)\\
    \midrule
    IP+CtrlN~~ &7.13E-04~~  &68.23~~ &93.56~~ &724.37 \\
    DisCo~~ &{3.78E-04}~~ &\bf{ 30.75}~~ &\underline{59.90}~~ &{292.80} \\
    MA~~ &\underline{ 3.13E-04}~~  &{32.09}~~ &{\bf 21.75}~~ &{\bf 179.07} \\
    Ours~~ &{\bf 3.06E-04}~~  &\underline{31.47}~~ &{63.26}~~ &\underline{286.33} \\
    \bottomrule
  \end{tabular}
  \caption{Quantitative comparisons on TikTok dataset.}
  \label{tab:comp:tiktok}
  \end{subtable}
  \begin{subtable}[t]{\linewidth}
  \centering
  \begin{tabular}{lcccc}
    \toprule
    \multirow{2}{*}{Method} &\multicolumn{2}{c}{Image} &\multicolumn{2}{c}{Video}\\
    \cmidrule(r){2-3} \cmidrule(r){4-5} & L1\(\downarrow\)  & FID\(\downarrow\) & FID-VID\(\downarrow\)& FVD\(\downarrow\)\\
    \midrule
    IP+CtrlN~~  &4.06E-04~~ &45.75~~ &38.48~~ &281.42 \\
    DisCo~~  &\underline{2.07E-04}~~ &{27.51}~~ &\underline{19.02}~~ &195.00 \\
    MA~~ &2.92E-04~~ &\underline{22.78}~~ &{\bf 19.00}~~ &{\bf 131.51} \\
    Ours~~ &{\bf 1.98E-04}~~  &{\bf 21.24}~~ &{20.15}~~ &\underline{168.02} \\
    \bottomrule
  \end{tabular}
  \caption{Quantitative comparisons on TED-talks dataset.}
  \label{tab:comp:tedtalks}
  \end{subtable}
  \caption{Quantitative performance comparison, with best performance in {\bf bold} and second best \underline{underlined}. MA corresponds to MagicAnimate (DensePose).}
  \label{tab:comp}
\end{table}

\subsection{Ablation Study}
\begin{figure}[!t]
  \centering
  \includegraphics[width=\linewidth]{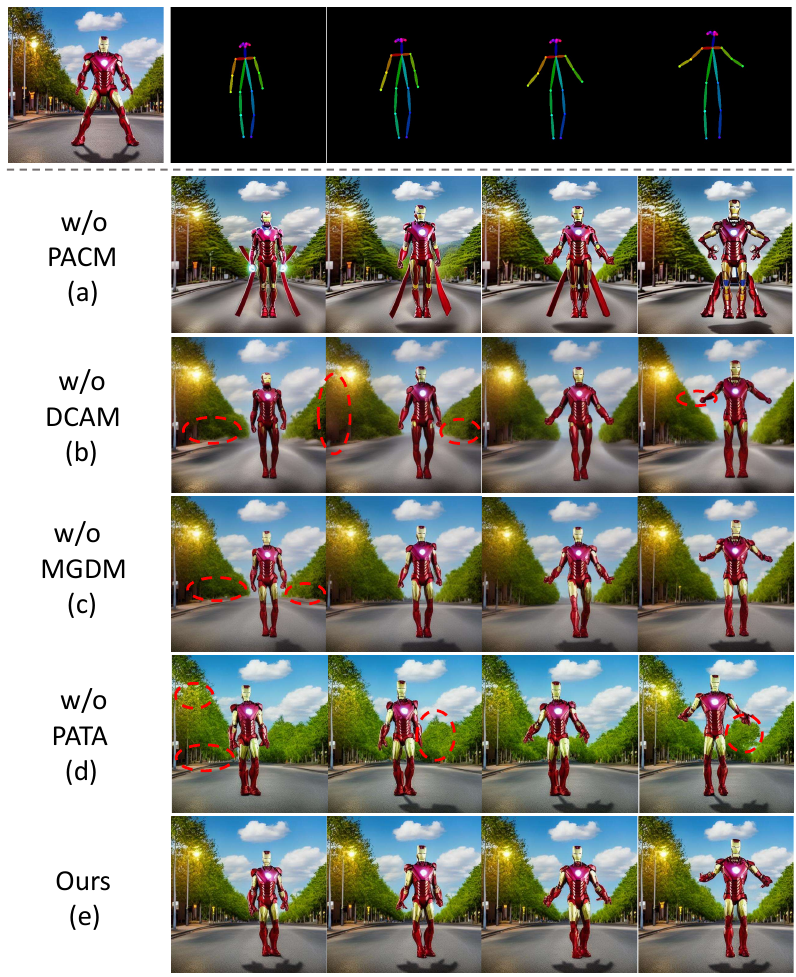}
  \caption{Visualization of ablation studies, with errors highlighted in red circles. Source prompt: ``An iron man on the road." 
  }
  \label{fig:abla}
\end{figure}
We conduct an ablation study to verify the effectiveness of each component of our framework and present the visualization results in Fig.~\ref{fig:abla}.
The leftmost one in the first row is the source image and the others are the target pose sequences. The following rows are generation results without certain components:
(a) Pose-Aware Control Module that effectively removes the interference of the source pose and maintains consistency of the content unrelated to the character; 
(b) Dual Consistency Attention Module that restores and preserves character identity while also improves temporal consistency; 
(c) Masked-Guided Decoupling Module that preserves fine-grained details and enhances animation fidelity; and 
(d) Pose Alignment Transition Algorithm that tackles the issue of pose misalignments while enabling smooth motion transitions.

\paragraph{PACM.}
Fig.~\ref{fig:abla}(a) illustrates the significant interference of the original pose on the generated actions. Due to the substantial difference between the posture of Iron Man's legs in the source and in the target, there is a severe breakdown in the leg area of the generated frame, undermining the generation of a reasonable target action. Moreover, character-independent scenes also have noticeable distortion.
\paragraph{DCAM.}
{From Fig.~\ref{fig:abla}(b) we can find that it fails to maintain character identity consistency without DCAM. And the missing pole and Iron Man's hand in the red circles reveal inter-frame inconsistency, indicating that both spatial and temporal consistency cannot be effectively maintained.
\paragraph{MGDM.}
Compared with our results in Fig.~\ref{fig:abla}(e), it can be observed that small signs are missing without MGDM. It shows that MGDM can effectively enhance the perception of fine-grained features and image fidelity.
\paragraph{PATA.}
Fig.~\ref{fig:abla}(d) verifies the proposed Pose Alignment Transition Algorithm. The red circles in the second frame indicate the spatial content misalignment. When Iron Man in the original image does not match with the input pose position, an extra tree appears in the original position of Iron Man. And such misalignment can also lead to disappearance of background details, e.g., streetlights and distant signage.

\section{Conclusion}
This paper proposes a novel zero-shot approach PoseAnimate to tackle the task of character animation for the first time. Through the integration of three key modules and an alignment transition algorithm, PoseAnimate can efficiently generate high-fidelity, pose-controllable and temproally coherent animations for a single image across diverse pose sequences. Extensive experiments demonstrate that PoseAnimate outperforms the state-of-the-art training based methods in terms of character consistency and detail fidelity. 

\section*{Acknowledgments}
This project was supported by NSFC under Grant No. 62032006.

\bibliographystyle{named}
\bibliography{ijcai24}

\end{document}